\documentclass[11pt,a4paper]{article}
\usepackage[hyperref]{acl2020}
\usepackage{times}
\usepackage{latexsym}
\usepackage{graphicx}
\usepackage{float}

\usepackage{microtype}

\aclfinalcopy 

\title{USR: An Unsupervised and Reference Free Evaluation Metric for Dialog Generation}

\author{Shikib Mehri and Maxine Eskenazi \\
  Dialog Research Center, Language Technologies Institute \\
  Carnegie Mellon University, USA \\
  \texttt{\{amehri,max\}@cs.cmu.edu}}

\date{}

\begin{document}
\maketitle
\begin{abstract}
The lack of meaningful automatic evaluation metrics for dialog has impeded open-domain dialog research. Standard language generation metrics have been shown to be ineffective for evaluating dialog models. To this end, this paper presents \textbf{USR}, an \textbf{U}n\textbf{S}upervised and \textbf{R}eference-free evaluation metric for dialog. USR is a reference-free metric that trains unsupervised models to measure several desirable qualities of dialog. USR is shown to strongly correlate with human judgment on both Topical-Chat (turn-level: \textbf{0.42}, system-level: \textbf{1.0}) and PersonaChat (turn-level: \textbf{0.48} and system-level: \textbf{1.0}). USR additionally produces interpretable measures for several desirable properties of dialog. 
\end{abstract}

\section{Introduction}

The lack of meaningful automatic evaluation metrics is a significant impediment for open-domain dialog generation research. Standard language generation metrics have been shown to be ineffective for dialog evaluation \citep{deriu2019survey,liu2016not}. Without well-accepted, meaningful automatic metrics, open-domain dialog researchers have come to rely on human evaluation. Due to its time- and cost-intensive nature, human evaluation is typically only used for the final dialog model. As such, during development dialog systems are generally optimized for poorly-correlated automatic metrics (e.g., F-1, BLEU, PPL) which can result in sub-par human evaluation scores \citep{dinan2019second}. To facilitate development of open-domain dialog models with meaningful automatic metrics, this paper presents the \textbf{U}n\textbf{S}upervised and \textbf{R}eference free (\textbf{USR}) evaluation metric for dialog. 

Standard automatic metrics for evaluating dialog generation (e.g., BLEU, F-1, METEOR, ROUGE) have several shortcomings that make them unsuitable for dialog evaluation: (1) The one-to-many nature of dialog \citep{zhao2017learning} makes word-overlap metrics ineffective for scoring valid system output that deviates from the ground-truth response \citep{liu2016not,gupta2019investigating}. (2) Human evaluation of dialog typically measures multiple properties (e.g., appropriate, interesting, consistent). Automatic metrics on the other hand, condense the multi-faceted nature of dialog quality to a single uninterpretable metric. (3) There are many definitions of what a \textit{good dialog} is and, as such, it is not feasible to construct a \textit{``one size fits all''} metric. Depending on the task and the data, the desired qualities of a dialog system may differ \cite{walker1997paradise,deriu2019survey}.

USR is a reference-free metric that consists of several interpretable sub-metrics which are combined in a configurable manner. Rather than relying on a ground-truth reference response, unsupervised models are trained to measure desired qualities of dialog (e.g., interesting, natural). As such, USR (1) alleviates the one-to-many issue of standard metrics, (2) produces interpretable measures for desirable properties of dialog, and (3) provides a configurable mechanism for combining several sub-metrics into an overall quality score.

To evaluate the performance of USR, human quality annotations were collected for models trained on the Topical-Chat \citep{gopalakrishnan2019topical} and the PersonaChat corpora \citep{zhang2018personalizing}. USR is shown to strongly correlate with human judgment on both Topical-Chat (turn-level Spearman: \textbf{0.42}, system-level Spearman: \textbf{1.0}) and PersonaChat (turn-level Spearman: \textbf{0.48} and system-level Spearman: \textbf{1.0}). The strong correlation with human judgment across two datasets and a variety of model types shows that USR is a valuable tool for the dialog community. Further, since USR does not require any explicit supervision, it has the potential to generalize to several dialog tasks and datasets.

The contributions of this paper as as follows: (1) a strongly-correlated, unsupervised and reference free metric is proposed for evaluating open-domain dialog systems, (2) a thorough human quality annotation is carried out and is released\footnote{\url{http://shikib.com/usr}} to facilitate future benchmarking of dialog evaluation metrics. 

\section{Related Work}

Standard automatic metrics for language generation correlate poorly with human judgement of dialog \citep{liu2016not,lowe-etal-2017-towards,gupta2019investigating}. For example, the F-1 score can be gamed by outputting the most frequent words, regardless of the context \citep{dinan2019second}. 

The poor performance of present metrics is largely due to the one-to-many nature of dialog \citep{zhao2017learning}. To avoid comparing to a single reference response, several authors have proposed using multiple reference responses. Multiple reference responses can be obtained with retrieval models \citep{galley-etal-2015-deltableu,Sordoni2015ANN} or through data collection \citep{gupta2019investigating}. These multi-reference metrics show improvement in performance, but it is infeasible to thoroughly cover the space of potential responses. As such, this paper addresses the one-to-many issue of dialog by presenting a reference-free metric.

\citet{lowe-etal-2017-towards} train ADEM to produce a quality score conditioned on the dialog context, the reference response and the generated response. \citet{venkatesh2018evaluating} present a framework for evaluation of Alexa prize conversations, which attains moderate correlation with user ratings. Both of these methods are trained on explicit quality annotations. In contrast, USR requires no explicit supervision and will more easily generalize to new datasets and tasks.

\citet{li2017adversarial} proposes a reference-free dialog evaluator which is trained to discriminate between human and generated responses. This work is similar to USR in that it evaluates the quality of a response without a reference or quality annotation training data. Using the evaluation model as a reward during reinforcement learning exhibited strong performance. However, correlation with human judgement was not evaluated. Intuitively, it appears insufficient to rely on a discriminator as a meaningful evaluation of dialog since this assumes that all human responses are perfect and all generated responses are imperfect.

\section{Human Quality Annotation}

To evaluate the correlation of automatic metrics with human judgment, human quality annotation was carried out across two open-domain dialog corpora. Generated responses were obtained from several models described in Section \ref{models}. For each dialog context, an additional human response was also written. Human annotation was then carried out on sixty dialog contexts, with six responses per context for Topical-Chat (four system outputs, one newly-annotated human output, one original ground-truth response) and five for PersonaChat (one less system output). Each response was given six different scores: Understandable (0-1), Natural (1-3), Maintains Context (1-3), Interesting (1-3), Uses Knowledge (0-1), Overall Quality (1-5). Three annotators labeled each response. 

The task instructions were very detailed in order to minimize subjectivity in the quality annotations. For example, individuals may differ in their definition of \textit{Interesting} (e.g., some individuals find football interesting, others do not). Thus, the instructions contained a clear, albeit somewhat rigid definition, of Interesting. The instructions for \textit{Overall Quality} annotation, however, were less rigid and therefore those annotations contain some amount of annotator-specific subjectivity. 

The data collection and experiments with PersonaChat were carried out to assess the generality of the USR metric. As such, the annotation questions used were specifically tailored to Topical-Chat, but are still suitable for PersonaChat.

\subsection{Topical-Chat Dataset}

The Topical-Chat dataset \citep{gopalakrishnan2019topical} is a large collection of human-human knowledge-grounded open-domain conversations that consists of 11,319 dialogs and 248,014 utterances. Following the same experimental setup as \citet{gopalakrishnan2019topical}, heuristics are employed to identify the most relevant fact for each response. As such, the task is to produce a response conditioned on both a dialog context and a fact. 

\subsection{PersonaChat Dataset}

The PersonaChat dataset \citep{zhang2018personalizing} is a corpus of human-human persona-conditioned conversations that consists of 10,907 dialogs and 162,064 utterances. Each worker is asked to condition their responses on a persona, which we consider to be analogous to the facts in the Topical-Chat corpus.

\subsection{Models}
\label{models}

\subsubsection{Topical-Chat Models}

A Transformer \citep{vaswani2017attention} is trained to produce the response, $r$, conditioned on dialog context, $c$, and fact, $f$. The input to the transformer is the concatenation of $c$ and $f$, similar to \citet{gopalakrishnan2019topical}. The transformer consists of 6 layers, a hidden size of 512, randomly-initialized word embeddings of size 300, a dropout rate of 0.1 and it is trained for 50 epochs.

\begin{figure}
    \centering
    \includegraphics[width=0.5\textwidth]{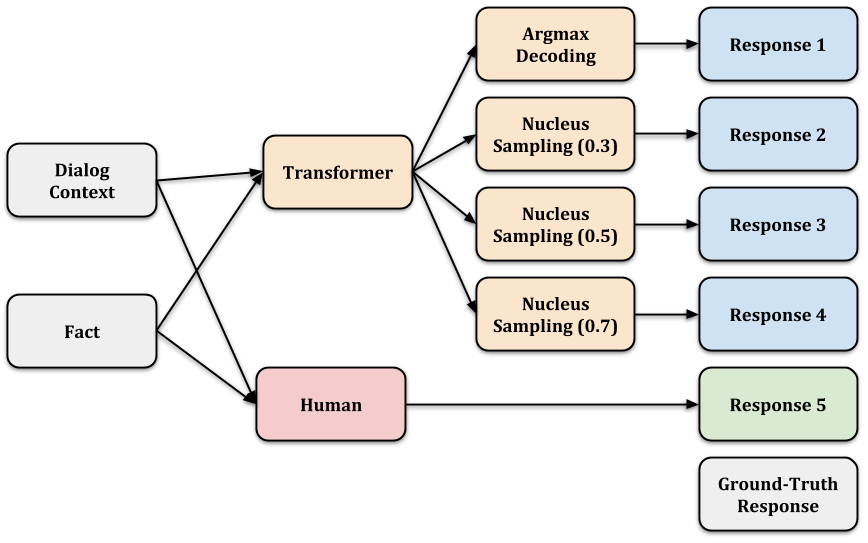}
    \caption{On the Topical-Chat corpus, six responses are obtained for each dialog context. Four use the trained Transformer model with different decoding strategies. One is a new human-generated response. One is the original ground-truth. A similar setup was employed for PersonaChat, albeit with different models.}
    \label{fig:responses}
\end{figure}

A single Transformer model is trained, which matches the automatic metrics reported by \citet{gopalakrishnan2019topical}. Different decoding strategies are used to obtain four different outputs from this model. In addition to standard argmax sampling, nucleus sampling \citep{holtzman2019curious} is used at three different rates: $p = \{0.3, 0.5, 0.7\}$. The outputs from these four decoding strategies are listed with the original ground-truth utterance and a new human-generated response, for a total of six responses for each context, as shown in Figure \ref{fig:responses}. 

\subsubsection{PersonaChat Models}

Three models were used to generate system outputs: a sequence-to-sequence model (Seq2Seq), an LSTM language model (LM) and a Key-Value Profile Memory Network (KV-MemNN). We use the pre-trained models provided in ParlAI\footnote{\url{https://github.com/facebookresearch/ParlAI/tree/master/projects/convai2}} for the ConvAI2 competition \citep{dinan2019second}.

A fourth open-source model was also used to produce output for quality annotation, however it was ultimately excluded from the released dataset and experiments due to possible data leakage. 

\subsection{Annotation}

Quality annotation was performed by six dialog researchers. Using a crowdsourcing platform, such as Amazon Mechanical Turk (AMT), would have allowed for more efficient and scalable annotation. However, crowdsourcing was not used because (1) the annotation instructions are lengthy, (2) a preliminary annotation pass was carried out, followed by a group discussion, (3) having many annotations from a few annotators allows examination of annotator-specific subjectivity. 

Annotators were provided with a set of instructions (Appendix A). A small preliminary annotation pass was carried out, with each individual annotating 5 dialog contexts (for a total of 30 responses). The inter-annotator agreement was computed for each of the questions. The instructions were refined after the preliminary pass and a discussion meeting (e.g., Maintains Context was changed to be a 3-point rating instead of a 2-point rating). After the instructions were modified,  the full annotation pass was carried out. 

Each response was rated according to the qualities mentioned at the beginning of this section. Instructions for each of qualities are summarized below:

\begin{itemize}
    \item \textit{Understandable (0 - 1):} Is the response understandable given the previous context?
    \item \textit{Natural (1 - 3):} Does the response seem to be something that a person would naturally say?
    \item \textit{Maintains Context (1 - 3):} Does the response serve as a valid continuation of the preceding conversation?
    \item \textit{Interesting (1 - 3):} Is the response dull or interesting?
    \item \textit{Uses Knowledge (0 - 1):} Given the fact that the response is conditioned on, how well does the response use that fact?
    \item \textit{Overall Quality (1 - 5):} Given your answers above, what is your overall impression of the quality of this utterance?
\end{itemize}

The instructions contained detailed descriptions and examples of what constitutes a response in each category (e.g., what makes a response score 2 on \textit{Maintains Context}). These instructions were written to minimize subjectivity in the annotations, which results in clear, agreed upon definitions. 

For Topical-Chat, the full annotation consisted of 60 dialog contexts randomly sampled from the \textit{frequent test set}, for a total of 360 responses scored on six different qualities. For PersonaChat, 60 dialog contexts were sampled from the ConvAI2 validation set, with a total of 300 responses scored on six different qualities. Each response was labeled by three different annotators. Annotators were randomly assigned to each dialog context.

\begin{table}
    \centering
    \begin{tabular}{|c|c|c|}
    \hline
    \textbf{Metric}  & \textbf{Spearman} & \textbf{Pearson} \\ \hline
    \multicolumn{3}{|c|}{Topical-Chat} \\ \hline
    Understandable & 0.5102 & 0.5102 \\
    Natural & 0.4871 & 0.4864 \\
    Maintains Context & 0.5599 & 0.5575 \\    
    Interesting & 0.5811 & 0.5754 \\   
    Uses Knowledge & 0.7090 & 0.7090 \\    
    Overall Quality & 0.7183 & 0.7096 \\    \hline
    \multicolumn{3}{|c|}{PersonaChat} \\ \hline
    Understandable & 0.2984 & 0.2984 \\
    Natural & 0.4842 & 0.4716 \\
    Maintains Context & 0.6125 & 0.6130 \\    
    Interesting & 0.4318 & 0.4288 \\   
    Uses Knowledge & 0.8115 & 0.8115 \\    
    Overall Quality & 0.6577 & 0.6603 \\    \hline
    \end{tabular}
    \caption{Inter-annotator agreement for all the metrics. For all the correlations presented in this table, $p <0.01$.}
    \label{tab:agreement}
\end{table}

\subsection{Analysis}

Inter-annotator agreements for the different ratings across both datasets are presented in Table \ref{tab:agreement}. The correlation between each pair of annotations is computed and the average correlation over all the pairs is reported. Correlation is used instead of Cohen's Kappa in order to better account for the ordinal nature of the ratings (i.e., $4$ should correlate better with $5$ than $1$), and to maintain consistency with the evaluation of the automatic metrics. Most inter-annotator correlations are above $0.4$, which indicates moderate to strong agreement. The low agreement for \textit{Understandable} on PersonaChat is likely a consequence of the simple language in the dataset. Most responses are understandable, except for those requiring background knowledge (e.g., that \textit{`cod'} is an acronym for \textit{`Call of Duty'}). Since the annotators have differing background knowledge, the few occasions where they fail to understand an utterance will differ, hence the lower agreement. The agreement for \textit{Overall Quality} is relatively high (0.71 for Topical-Chat and 0.66 for PersonaChat) which suggests that any ambiguity in the specific dialog qualities is mitigated when the annotator is asked for an overall impression.

\begin{table*}
    \centering
    \begin{tabular}{|c|c|c|c|c|c|c|}
    \hline
    \textbf{System} & \textbf{Und} (0-1) & \textbf{Nat} (1-3) & \textbf{MCtx} (1-3) & \textbf{Int} (1-3) & \textbf{UK} (0-1) & \textbf{OQ} (1-5) \\ \hline
    
    \multicolumn{7}{|c|}{Topical-Chat} \\ \hline
    Original Ground-Truth    &  0.95  & 2.72 & 2.72 & 2.64 & 0.72 & 4.25 \\ 
    Argmax Decoding &  0.60  & 2.08 & 2.13 & 1.94 & 0.47 & 2.76 \\
    Nucleus Sampling (0.3) & 0.51 & 2.02 & 1.90 & 1.82 & 0.42 & 2.40 \\
    Nucleus Sampling (0.5) & 0.48 & 1.92 & 1.93 & 1.72 & 0.34 & 2.29 \\
    Nucleus Sampling (0.7) & 0.52 & 2.01 & 1.87 & 1.80 & 0.37 & 2.39 \\
    New Human Generated & \textbf{0.99} & \textbf{2.92} & \textbf{2.93} & \textbf{2.90} & \textbf{0.96} & \textbf{4.80} \\ \hline    \multicolumn{7}{|c|}{PersonaChat} \\ \hline
    Original Ground-Truth    &  0.99 & 2.89 & 2.82 & 2.67 & 0.56 & 4.36 \\ 
    Language Model & 0.97 & 2.63 & 2.02 & 2.24 & 0.08 & 2.98 \\
    LSTM Seq2Seq & 0.92 & 2.64 & 2.49 & 2.29 & 0.47 & 3.47 \\
    KV-MemNN &  0.93  & 2.70 & 2.18 & 2.56 & 0.17 & 3.25 \\
    New Human Generated & \textbf{1.00} & \textbf{2.97} & \textbf{2.88} & \textbf{2.87} & \textbf{0.96} & \textbf{4.80} \\ \hline   

    \end{tabular}
    \caption{Average scores for the six different responses, on the six quality: Understandable, Natural, Maintains Context, Interesting, Uses Knowledge and Overall Quality.}
    \label{tab:human_scores}
\end{table*}

Table \ref{tab:human_scores} presents the scores for the different systems on each of the six qualities. Across both datasets and all qualities, the new human generated response strongly outperforms all other response types, even the original ground truth. This may be because the new human generated response was written with this quality annotation in mind, and as such is optimized for turn-level evaluation. On the other hand, the workers who produced the original ground-truth response, were more concerned with the quality of the overall dialog than with the quality of each individual response. 

On the Topical-Chat corpus, argmax decoding has a moderately higher performance over the nucleus sampling \citep{holtzman2019curious} methods. This should not be taken as an indication that argmax decoding is the superior method, since the hyperparameters (e.g., temperature) were not tuned for nucleus sampling. It should be noted that the objective was not to train and evaluate the best performing models, but instead to produce responses of varying qualities and obtain accurate human judgements of these responses.

\begin{figure}
    \centering
    \includegraphics[width=0.5\textwidth]{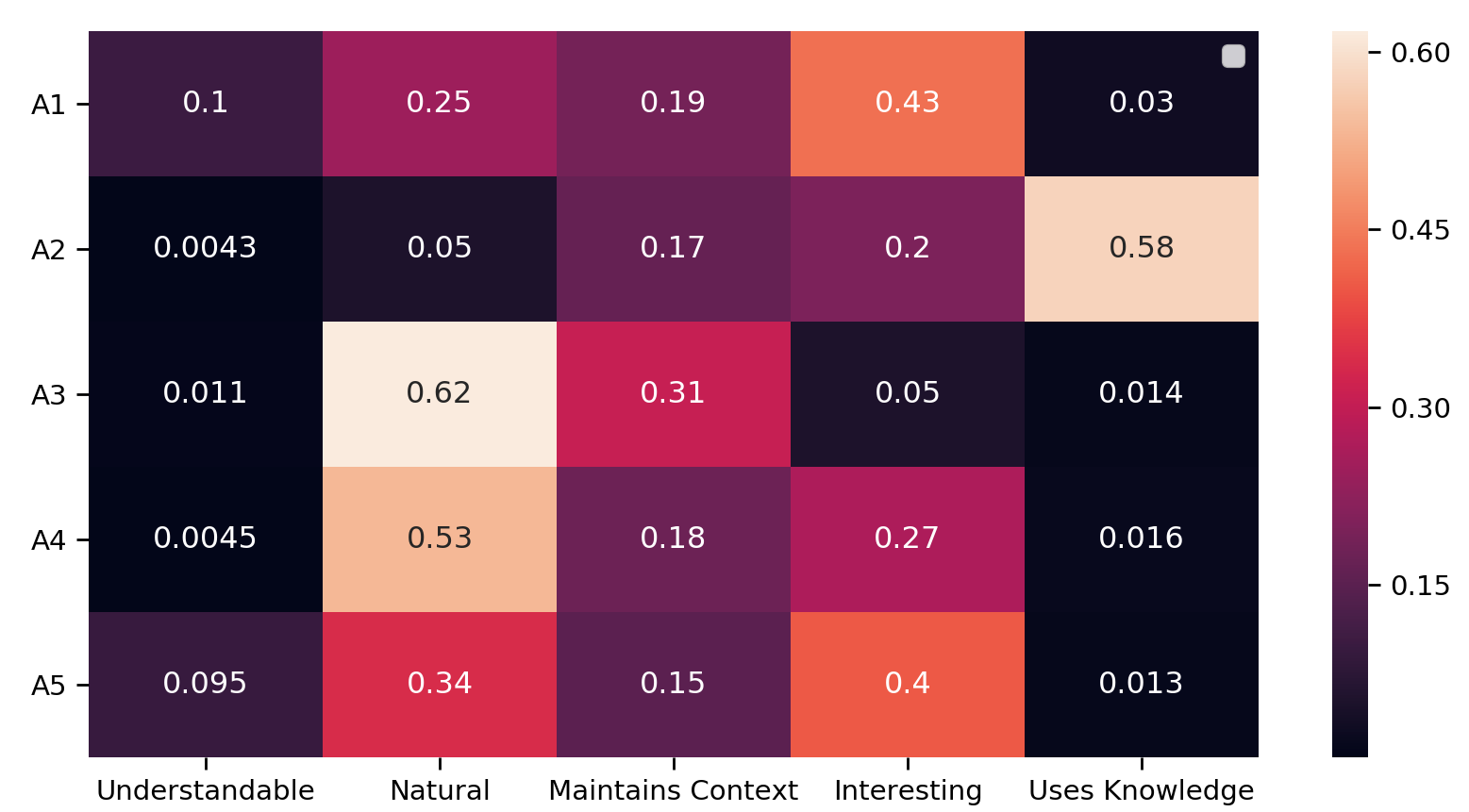}
    \caption{Weight attributed to each of the five specific metrics by each annotator, when labeling \textit{Overall Quality}. Lighter colors signify more weight.}
    \label{fig:weights}
\end{figure}

A regression was trained to map from the five ratings to the overall score in order to analyze the relationship between them. For better interpretability of the regression weights, the scores were normalized (using z-score) before training the regression. For better interpretability, a softmax was computed over the weights. Since individuals may differ in their definition of a \textit{good response}, a specific regression is trained for each of the five annotators who labeled responses for the Topical-Chat corpus.  Figure \ref{fig:weights} displays the weights attributed to each of the five qualities by each of the annotators.

Annotators attributed different weights to the specific features. For example, A3 emphasized naturalness while A2 paid more attention to whether a response was grounded on knowledge. Despite the differences across annotators, a good response was generally expected to be natural, maintain context, and be interesting. These annotator-specific weights demonstrate that individuals define \textit{good dialog} differently. Future work could explore personalized dialog evaluation wherein the evaluation metric is tailored to a specific individual.

A potential criticism of this quality annotation could be that certain dialog qualities are missing. To address concerns about the \textit{completeness} of the set of five qualities, a regression can be trained to produce the overall score conditioned on the quality ratings. The Spearman correlation between the predicted score and the original overall score is \textbf{0.9654}, which signifies that the set of qualities is thorough and contains enough information to reflect the overall quality of the response.  
\label{analysis}

\section{Automatic Metrics}
\label{am}
This section describes the automatic metrics explored for evaluating generated responses. Section \ref{baseline} describes several existing metrics that were studied. Section \ref{new} presents USR, a novel unsupervised and reference-free metric. 
\subsection{Baseline Metrics}
\label{baseline}

Several existing and easily-applicable metrics for dialog evaluation are compared. the list of available metrics is not exhaustive. Only the most commonly used and the most accessible are addressed.

\textbf{F-1} score computes the word-overlap between the generated response and the ground-truth, by taking the harmonic mean of the precision and recall. It is one of the four metrics used by the creators of the Topical-Chat dataset \citep{gopalakrishnan2019topical}, along with perplexity and unique unigram/bigram counts. \citet{dinan2019second} described a simple adversarial example that attains a high F-1 score on PersonaChat. We produce a similar example for the Topical-Chat dataset and find that always outputting a concatenation of the ten most common tokens in the dataset (\textit{``. i the , that a to it is of''}) attains an F-1 score of \textbf{25.6} which is a \textbf{+3.6} improvement over the Transformer presented by \citet{gopalakrishnan2019topical}.

\textbf{BLEU} \citep{papineni2002bleu} is a well-known word overlap metric that computes n-gram precision between the generated sequence and the reference. Because precision favors shorter sentences, BLEU also adds a brevity penalty that punishes shorter sentences. BLEU has been found to correlate poorly with human judgment \citep{liu2016not,lowe-etal-2017-towards,gupta2019investigating}.

\textbf{METEOR} \citep{denkowski2014meteor} was designed as an improvement on BLEU using a harmonic mean of precision and recall, as well as stemming and synonyms.  

\textbf{ROUGE-L} \citep{lin2004rouge} identifies the longest common subsequence between the generated and reference sequence to better account for sentence-level structure when computing word overlap.

\textbf{Greedy Matching} \citep{rus2012comparison} is an embedding-based metric that greedily matches each word in the generated sequence to a reference word based on the cosine similarity of their embeddings. The final score is then an average over all the words in the generated sequence. 

\textbf{Embedding Average} \citep{wieting2015towards} computes a \textit{sentence embedding} for both the generated sequence and the ground-truth response by taking an average of word embeddings. The score is then a cosine similarity of the average embedding for both the generated and reference sequence.

\textbf{Vector Extrema} \citep{forgues2014bootstrapping} follows a similar setup to Embedding Average, where the score is the cosine similarity between sentence embeddings. Rather than taking an average over word embeddings, this method identifies the \textit{maximum} value for each dimension of the word embedding. Taking the maximum is motivated by the idea that common words will be de-emphasized as they will be closer to the origin. Vector Extrema has been shown to perform better on dialog tasks than other metrics \citep{gupta2019investigating,liu2016not}. 

\textbf{Skip-Thought} \citep{kiros2015skip} uses a recurrent neural network to produce a sentence-level embedding for the generated and reference sequences. A cosine similarity is then computed between the two embeddings. The implementation provided by \citet{sharma2017relevance} is used.

\textbf{BERTScore} \citep{zhang2019bertscore} uses a pre-trained BERT \citep{devlin2018bert} model to greedily match each word in a reference response with one word in the generated sequence. By doing so, it computes the recall of the generated sequence. BERTScore was shown to have strong system-level and segment-level correlation with human judgment on several machine translation and captioning tasks. However, although it is a more sophisticated metric, it still compares word similarity between a reference and a generated sequence. While this method may work well for tasks where there is a limited space of outputs for each input (e.g., captioning, translation), it is ineffective at dealing with the one-to-many nature of dialog.

\subsection{Proposed Metric}
\label{new}

This section proposes describes the USR metric, an unsupervised, reference-free evaluation metric for dialog. USR leverages pre-trained language models, specifically RoBERTa \citep{liu2019roberta}, to measure properties of dialog. USR is designed to be reference-free because there is no \textit{one right answer} due to the inherent one-to-many nature of dialog \citep{zhao2017learning}. 

Several sub-metrics were developed for the different qualities of dialog (e.g., Natural, Interesting, Uses Knowledge). While USR measures the overall quality of a response, its sub-metrics assess specific dialog qualities and therefore facilitate better understanding of a model's performance.

\subsubsection{Masked Language Modelling Metric}
\label{mlm}

The masked language modelling (MLM) metric uses a fine-tuned RoBERTa \citep{liu2019roberta} model to estimate the likelihood of a response. RoBERTa is pre-trained on a massive amount of English data and fine-tuned on the corpus being evaluated (either Topical-Chat or PersonaChat), making it capable of identifying unnatural and incorrect responses. The likelihood estimated by the fine-tuned RoBERTa model is used as an automatic metric for evaluating the understandability and naturalness of responses.

The RoBERTa-base model \citep{liu2019roberta} is fine-tuned on the training set of the Topical-Chat corpus \citep{gopalakrishnan2019topical} using the implementation open-sourced by \citet{Wolf2019HuggingFacesTS}. The language model is fine-tuned on only the dialog, without any of the facts, for a single epoch.

\begin{figure}
    \centering
    \includegraphics[width=0.5\textwidth]{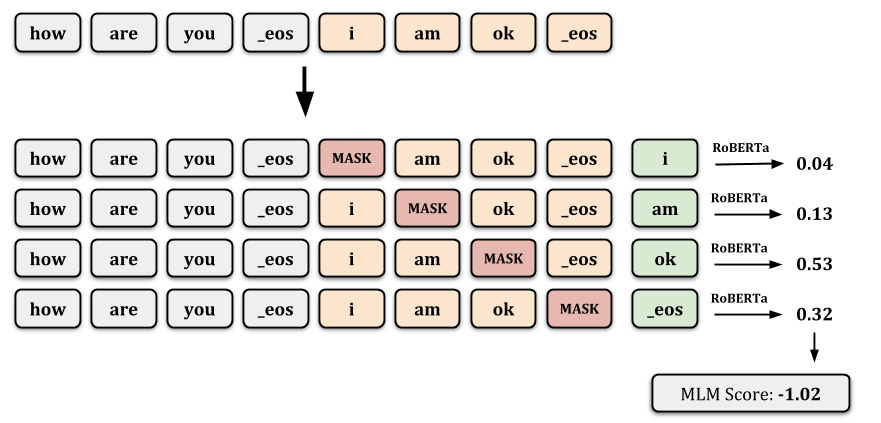}
    \caption{Visualization of the masked language modelling (MLM) metric. Context words are in grey; response words are in red. The red words are masked, and RoBERTa must predict the likelihood of their true value (shown in green).}
    \label{fig:mlm_score}
\end{figure}

RoBERTa uses both past and future context to predict a probability distribution for a masked word. The input sequence to MLM is a concatenation of a dialog context, $c$, and a response, $r$. One word at a time, each word in $r$ is masked and its log likelihood is computed. Given the masked log-likelihood for the $i$-th word of $r$ as $l_i$, the value of the metric is then computed to be $-\sum_{i}^{|r|} l_i$. Figure \ref{fig:mlm_score} visualizes this process.

\subsubsection{Dialog Retrieval Metrics}

Recent research has highlighted the complementary nature of dialog retrieval and generation with respect to multi-tasking \citep{wolf2019transfertransfo} and pre-training \citep{mehri2019pretraining}. Because of this complimentary nature, using dialog retrieval (DR) for \textit{evaluating} generative models is an intuitive choice, especially for metrics like \textit{Maintains Context} and \textit{Uses Knowledge}.

The fine-tuned RoBERTa model described in Section \ref{mlm} is further fine-tuned for the retrieval task. This task is set up in the same manner as the Ubuntu dialog corpus \citep{lowe2015ubuntu}. The model is trained given a context $x$, a response $r$, and a binary label $y$ indicating whether $r$ is the true response or randomly sampled. The context $x$ may consist of the dialog history and the fact, denoted $c$, or just the fact, denoted $f$. Two different versions of the dialog retrieval (DR) metric are trained, with different values of $x$. The DR metric score is defined to be the probability $P(y = 1 | ~x, r)$ a given DR metric model produces. 

Though the DR metric is trained for the task of retrieval, this is done in an unsupervised manner. The retrieval task is an unsupervised task since it requires no additional labels during training (e.g., explicit quality annotations).

The DR metric is appropriate for \textit{Maintains Context}, \textit{Interesting} and \textit{Uses Knowledge}. If a retrieval model predicts that a generated response is contextually relevant to a dialog context, it indicates that the response \textit{Maintains Context}. Likewise, if a retrieval model predicts that the response $r$ is contextually relevant to fact $f$, it signifies that $r$ most likely \textit{Uses Knowledge}. 

\textit{Interesting} is the measure of whether the response is dull/generic or if it provides some interesting/engaging information. The DR metric is trained to distinguish between a ground-truth response ($y=1$) and a randomly sampled response ($y=0$). Generic responses are applicable to many contexts, and will often appear as both ground-truth responses and randomly sampled responses. As such, the model will likely learn to assign a low probability distribution to these generic responses and will often output $P(y=1 | ~r, x) = 0.5$. As such, generic responses will generally be scored lower than other contextually relevant, interesting responses. The DR metrics will learn to favor responses that are unique to a given context $x$, rather than being applicable to many different contexts.
\label{dr}

\subsubsection{The USR Metric}

Given meaningful automatic metrics for each of the five dialog qualities, USR combines the scores into an overall measure that correlates well with \textit{Overall Quality} ratings. 

In Section \ref{analysis}, a regression model was trained to reproduce the overall score from each of the specific quality scores. The predictions of this regression model attained a 0.9654 Spearman correlation with the original scores. This same regression is used by USR on top of the automatic metrics presented in Sections \ref{mlm} and \ref{dr}. 

USR combines its sub-metrics into one measure of overall quality. This combination is configurable, adaptable to different datasets or tasks. For example, if a specific application prefers natural responses over interesting ones, the weights of the regression model can be adjusted. Analysis demonstrated that individuals used different weights when producing the overall score (Figure \ref{fig:weights}). USR might be able to be personalized for specific individuals by adjusting the weights of the regression model.



\section{Results}

\begin{table}
\centering
\begin{tabular}{|l|c|c|}
\hline
Metric                    & Spearman & Pearson  \\\hline
\multicolumn{3}{|c|}{Understandable} \\ \hline
BERTScore (base)          & 0.2502   & 0.2611   \\
USR - MLM                       & \textbf{0.3268}   & \textbf{0.3264}   \\
USR                       & 0.3152   & 0.2932   \\
\hline
\multicolumn{3}{|c|}{Natural} \\ \hline
BERTScore (base)          & 0.2094   & 0.2260   \\
USR - MLM                       & \textbf{0.3254}   & \textbf{0.3370}  \\
USR                       & 0.3037   & 0.2763   \\\hline
\multicolumn{3}{|c|}{Maintains Context} \\ \hline
METEOR                    & 0.3018   & 0.2495   \\
USR - DR (x = c)   & 0.3650  & 0.3391   \\
USR                       & \textbf{0.3769}   & \textbf{0.4160}   \\\hline
\multicolumn{3}{|c|}{Interesting} \\ \hline
BERTScore (base)          & 0.4121   & 0.3901  \\
USR - DR (x = c)   & \textbf{0.4877}   &  0.3533   \\
USR                       & 0.4645   & \textbf{0.4555}   \\\hline
\multicolumn{3}{|c|}{Uses Knowledge} \\ \hline
METEOR                    & 0.3909  & \textbf{0.3328}   \\
USR - DR (x = f)   & \textbf{0.4468}   &  0.2220   \\
USR                       & 0.3353   & 0.3175   \\\hline

\end{tabular}
\caption{Turn-level correlations on Topical-Chat. We show: (1) best non-USR metric, (2) best USR sub-metric and (3) USR metric. All measures in this table are statistically significant to $p < 0.01$.}
\label{tab:all_turn}
\end{table}

\begin{table}
\centering
\begin{tabular}{|l|c|c|}
\hline
Metric                    & Spearman & Pearson  \\\hline
\multicolumn{3}{|c|}{Understandable} \\ \hline
BERTScore (base)          & \textit{0.0685}   & \textit{0.0672}   \\
USR - MLM               & \textit{0.1186}   & \textbf{0.1313}   \\
USR                       & \textbf{0.1324}   & \textit{0.1241}   \\
\hline
\multicolumn{3}{|c|}{Natural} \\ \hline
VectorExtrema             & 0.1375   & 0.1458   \\
USR - DR (x = c)          & 0.2291   & 0.1733  \\
USR                       & \textbf{0.2430}   & \textbf{0.1862}   \\\hline
\multicolumn{3}{|c|}{Maintains Context} \\ \hline
METEOR                    & 0.2564   & 0.2500   \\
USR - DR (x = c)   & \textbf{0.5625}  & 0.6021   \\
USR                       & 0.5280   & \textbf{0.6065}   \\\hline
\multicolumn{3}{|c|}{Interesting} \\ \hline
BERTScore (base)          & \textit{0.0491}  & \textit{0.0325}  \\
USR - DR (x = c)   & \textbf{0.2634}   &  \textit{0.0606}   \\
USR                       & \textit{0.0171}   & \textit{0.0315}   \\\hline
\multicolumn{3}{|c|}{Uses Knowledge} \\ \hline
METEOR                    & 0.1719  & 0.1678   \\
USR - DR (x = c)   & \textbf{0.6309}   &  \textbf{0.4508}   \\
USR                       & 0.3177   & 0.4027   \\\hline

\end{tabular}
\caption{Turn-level correlations on Persona-Chat. We show: (1) best non-USR metric, (2) best USR sub-metric and (3) USR metric. All values with $p > 0.05$ are italicized.}
\label{tab:all_turn_pc}
\end{table}

\begin{table}
\centering
\begin{tabular}{|l|c|c|}
\hline
Metric                    & Spearman & Pearson  \\\hline
\multicolumn{3}{|c|}{Word-Overlap Metrics} \\ \hline
F-1                       & 0.1645   & 0.1690   \\
BLEU-1                    & 0.2728   & 0.2876   \\
BLEU-2                    & 0.2862   & 0.3012   \\
BLEU-3                    & 0.2569   & 0.3006   \\
BLEU-4                    & 0.2160   & 0.2956   \\
METEOR                    & 0.3365   & 0.3908   \\
ROUGE-L                   & 0.2745   & 0.2870   \\\hline
\multicolumn{3}{|c|}{Embedding Based Metrics} \\ \hline
Greedy Matching           & 0.1712   & 0.1943   \\
Embedding Average         & 0.1803   & 0.2038   \\
Vector Extrema            & 0.2032   & 0.2091   \\
Skip-Thought              & \textit{0.1040} & \textit{0.1181} \\
BERTScore (base)  & 0.3229   & 0.3540   \\
BERTScore (large) & 0.2982   & 0.3252   \\ \hline
\multicolumn{3}{|c|}{Reference Free Metrics} \\ \hline
USR - MLM               & 0.3086   & 0.3345   \\
USR - DR (x = c)   & 0.3245   & 0.4068   \\
USR - DR (x = f)   & 0.1419   & 0.3221   \\
USR             & \textbf{0.4192 }  & \textbf{0.4220 } \\\hline
\end{tabular}
\caption{Turn-level correlations between all automatic metrics and the \textit{Overall Quality} ratings for the Topical-Chat corpus. All values with $p > 0.05$ are italicized.}
\label{tab:turn_overall}
\end{table}

\begin{table}
\centering
\begin{tabular}{|l|c|c|}
\hline
Metric                    & Spearman & Pearson  \\\hline
\multicolumn{3}{|c|}{Word-Overlap Metrics} \\ \hline
F-1 & 0.1422 & \textit{0.1241} \\
BLEU-1 & \textit{0.0434} & \textit{0.0469} \\
BLEU-2 & \textit{0.1122} & \textit{0.0943} \\
BLEU-3 & \textit{0.1202} & \textit{0.0924} \\
BLEU-4 & 0.1353 & \textit{0.0899} \\
METEOR & 0.2527 & 0.2713 \\
ROUGE-L & \textit{0.0659} & \textit{0.0385} \\ \hline
\multicolumn{3}{|c|}{Embedding Based Metrics} \\ \hline
Greedy Matching & \textit{0.0916} & \textit{0.0625} \\
Embedding Average & \textit{0.1182} & 0.1428 \\
Vector Extrema & 0.1570 & 0.1410 \\
Skip-Thought & \textit{-0.0393} & \textit{-0.0452} \\
BERTScore (base) & 0.1690 & 0.1526 \\
BERTScore (large) & 0.1518 & \textit{0.1211} \\ \hline
\multicolumn{3}{|c|}{Reference Free Metrics} \\ \hline
USR-MLM & \textit{0.0795} & \textit{0.0788} \\
USR-DR (x = f) & \textit{-0.0495} & \textit{-0.0454} \\
USR-DR (x = c) & \textbf{0.4814} & \textbf{0.6087} \\
USR & 0.4693 & 0.4115 \\ \hline
\end{tabular}
\caption{Turn-level correlations between all automatic metrics and the \textit{Overall Quality} ratings for the PersonaChat corpus. All values with $p > 0.05$ are italicized.}
\label{tab:turn_overall_pc}
\end{table}

This section evaluates all of the automatic metrics described in Section \ref{am}, by comparing them to human judgement. The best sub-metrics for each dialog quality are used as input for the regression model of the USR metric. While the best performing sub-metrics are not consistent across the two datasets, the USR metric nonetheless exhibits strong results. The annotations for the original ground-truth are not used for evaluation, in order to accurately compare referenced and reference-free metrics.

Table \ref{tab:all_turn} shows turn-level correlations of the best automatic metrics for each dialog quality on Topical-Chat. USR is shown to strongly outperform both word-overlap and embedding-based metrics across all of the dialog qualities. Interestingly, the best non-USR metric is consistently either METEOR or BERTScore -- possibly because both methods are adept at comparing synonyms during evaluation. For some dialog qualities, the overall USR metric outperforms the best sub-metric. For example, USR does better for \textit{Maintains Context} than USR-DR. This is likely because the information from the other sub-metrics (e.g., \textit{Uses Knowledge}) is valuable and effectively leveraged by USR.

Table \ref{tab:all_turn_pc} reports the turn-level correlations of the best automatic metrics for each dialog quality on the PersonaChat corpus. Across all dialog qualities, USR strongly outperforms the word-overlap and embedding-based metrics. Conversations in PersonaChat generally consist of individuals communicating facts from their own persona in a relevant and coherent manner. As such, when models trained on PersonaChat produce subpar outputs, it is generally because the outputs either (1) do not effectively use the persona or (2) are not relevant/coherent to the dialog context. This explains why the correlations are significantly higher for \textit{Maintains Context} and \textit{Uses Knowledge}. As a consequence of PersonaChat's strong dependency on both the dialog context and the persona, USR-DR (x = c) which uses both the dialog context and the persona to perform dialog retrieval, generally outperforms all other metrics.  

Table \ref{tab:turn_overall} shows turn-level correlation with the \textit{Overall Quality} ratings on Topical-Chat, for all of the automatic metrics. USR shows a strong improvement over all other methods. This strong performance can be attributed to two factors: (1) the ability of MLM and DR to accurately quantify qualities of a generated response without a reference response, and (2) the ability of USR to effectively combine MLM and DR into a better correlated overall metric. 

USR shows a similar improvement over all other metrics on PersonaChat, as shown in Table \ref{tab:turn_overall_pc}. However, DR (x = c) outperforms USR despite the fact that four out of the five sub-metrics input into the USR regression are DR (x = c). This result is probably due to PersonaChat's strong dependancy on both dialog context and persona, both of which DR (x = c) explicitly leverages.

We compute the system-level correlation between all automatic metrics and the \textit{Overall Quality} ratings. USR significantly ($p < 0.01$) outperforms all other metrics with a Spearman correlation of \textbf{1.0} on both datasets and Pearson correlations of \textbf{0.92} (Topical-Chat) and \textbf{0.82} (PersonaChat). The full set of system-level correlations can be found in the appendix.

These results demonstrate USR's effectiveness. It strongly outperforms other metrics on both turn-level and system-level correlations. \citet{gopalakrishnan2019topical} use the F-1 score as their primary automatic evaluation metric when presenting Topical-Chat. The results demonstrate a significant difference between USR and the F-1 score, suggesting that USR is a better metric for the Topical-Chat corpus.




\section{Discussion}
USR achieves statistically significant correlations with human judgement. The results hold across two datasets, Topical-Chat \citep{gopalakrishnan2019topical} and PersonaChat \citep{zhang2018personalizing}. 

USR is configurable. Notably it is composed of several specific dialog quality sub-metrics. These sub-metrics are combined in a configurable manner, using a regression. For other tasks, datasets or even users, this regression can be adjusted, allowing qualities to be removed or re-weighted. Additional sub-metrics could be added.

USR should be used alongside human evaluation. USR was created to facilitate development and tuning of dialog models. As such, USR can be used for model selection and hyperparameter tuning. USR should not be used to claim superior performance over another method.

USR may not work with non-generative models, which were not addressed here. Responses produced by a model that is too similar to the evaluation models (e.g., to DR) are a particular concern.

\section{Conclusions}

This paper presents \textbf{USR}, an \textbf{U}n\textbf{S}upervised and \textbf{R}eference-free evaluation metric for dialog. To address the shortcomings of standard metrics for language generation, USR (1) is reference-free, (2) is composed of multiple sub-metrics that evaluate specific qualities of dialog, (3) has a definition of \textit{good dialog} that is configurable. 
Thus the metric may be adapted to different tasks and datasets. 
USR is shown to strongly correlate with human judgment on Topical-Chat (turn-level: \textbf{0.42}, system-level: \textbf{1.0}) and PersonaChat  (turn-level: \textbf{0.48}, system-level: \textbf{1.0}).

\section{Acknowledgements}

We thank the following individuals for their help with annotation: Evgeniia Razumovskaia, Felix Labelle, Mckenna Brown and Yulan Feng.



\bibliography{acl2020}
\bibliographystyle{acl_natbib}

\clearpage
\appendix
\section{Annotation Instructions}

Tables 6, 7 and 8 show the annotation instructions used for human quality annotation. These instructions and examples are verbatim what was shown to the annotators. 

\begin{table*}
    \begin{tabular}{|c|p{0.7\linewidth}|}
        \hline
        \multicolumn{2}{|c|}{\textbf{Annotation Instructions}} \\ \hline
        \multicolumn{2}{|p{\linewidth}|}{You will be given a conversation between two individuals. You will then be given several potential responses for the next turn in the conversation. These responses all concern an interesting fact, which will be provided as well.} \\
        \multicolumn{2}{|p{\linewidth}|}{Your task is to rate each of the responses on several metrics. The response for one metric should not influence the other metrics. For example, if a response is not understandable or has grammatical errors -- you should try to ignore this when considering whether it maintains context or if it is interesting.} \\
        \multicolumn{2}{|p{\linewidth}|}{\textbf{Please make sure you read and understand these instructions carefully.} Feel free to ask if you require clarification. Please keep this document open while reviewing, and refer to it as needed.} \\ \hline
        Understandable (0-1) &  Is the response understandable in the context of the history? (Not if it’s on topic, but for example if it uses pronouns they should make sense) 
        \begin{itemize}
            \item A score of 0 (no) means that the response is difficult to understand. You do not know what the person is trying to say. \begin{itemize}
                \item \textit{i did n't know that . i love to watch the movie inception , it 's also the first racing movie to be a woman haha . i guess the movie was originally titled " inception " awesome movie !}
                \item \textbf{Context:} in my religion , there is no star . how about you \textbf{Response:} \textit{yeah it was back in 1975 .}
            \end{itemize}
            \item A score of 1 (yes) means that the response is understandable. You know what the person is trying to say. \begin{itemize}
                \item \textit{my favorite role would have to be quarterback . it is such an interesting role .}
                \item \textit{that is true . i think lebron is the highest paid celebrity , i wonder if he will be in the space jam sequel .}
            \end{itemize}
        \end{itemize} \\ \hline
        Natural (1-3) &  Is the response naturally written?
        \begin{itemize}
            \item A score of 1 (bad) means that the response is unnatural. \begin{itemize}
                \item \textbf{Context:} A: wow . do you believe in stars of the zodiac ? what is your star ? B: in my religion , there is no star . how about you \textbf{Response:} \textit{yeah , it was back in 1975 .}
                \item \textit{i think he is , he is a great teacher and he also taught ellie kemper , she is a great teacher}

            \end{itemize}
            \item A score of 2 (ok) means the response is strange, but not entirely unnatural. \begin{itemize}
                \item \textbf{Context:} A: wow . do you believe in stars of the zodiac ? what is your star ? B: in my religion , there is no star . how about you \textbf{Response:} \textit{i read it sometimes for the fun of it .}
            \end{itemize}
            \item A score of 3 (good) means that the response is natural.  \begin{itemize}
                \item \textit{i think it 's funny that the soviet union sent a spacecraft to venus .}
            \end{itemize}            
        \end{itemize} \\ \hline

    \end{tabular}
    \caption{Annotation instructions (part 1 of 3).}
\end{table*}

\begin{table*}
    \begin{tabular}{|p{0.1\linewidth}|p{0.9\linewidth}|}
        \hline
        \multicolumn{2}{|c|}{\textbf{Annotation Instructions (ctd.)}} \\ \hline
  
        Maintains Context (1-3) &  Does the response serve as a valid continuation of the conversation history?
        \begin{itemize}
            \item A score of 1 (no) means that the response drastically changes topic or ignores the conversation history. \begin{itemize}
                \item \textbf{Context:} A: wow . do you believe in stars of the zodiac ? what is your star ? B: in my religion , there is no star . how about you \textbf{Response:} \textit{i think it 's funny that the soviet union sent a spacecraft to venus .}
            \end{itemize}
            \item A score of 2 (somewhat) means the response refers to the conversation history in a limited capacity (e.g., in a generic way) and shifts the conversation topic. \begin{itemize}
                \item \textbf{Context:} i do like some drama stuff , yeah he was awesome in that . \textbf{Response:} \textit{yeah . do you like jon hamm ?}
                \item \textbf{Context:}  i believe that ! he would have played longer i 'm sure if he did the granny style approach to shooting freethrows ! \textbf{Response:} \textit{i agree . did you know that space jam is the highest grossing basketball movie of all time ?}
            \end{itemize}
            \item A score of 3 (yes) means the response is on topic and strongly acknowledges the conversation history.  \begin{itemize}
                \item \textbf{Context:}  B: wow , that 's great . especially because more than of nba players go broke 5 years after retirement . A: i believe that ! he would have played longer i 'm sure if he did the granny style approach to shooting freethrows ! \textbf{Response:} \textit{a lot of players can make money by starring in movies . did you know space jam is the highest grossing movie of all time ? maybe one of the broke retired players can be in the sequel !}
                \item \textbf{Context:} B: you like drama ? patrick stewart teaches classes now . i loved him in star trek A: i do like some drama stuff , yeah he was awesome in that . \textbf{Response:} \textit{jonn hamm was also a drama teacher . he taught erin from the office}
            \end{itemize}            
        \end{itemize} \\ \hline

        Interesting (1-3) &  Is the response dull/interesting?
        \begin{itemize}
            \item A score of 1 (dull) means that the response is generic and dull. \begin{itemize}
                \item \textit{that’s true . i agree .}
                \end{itemize}
            \item A score of 2 (somewhat interesting) means the response is somewhat interesting and could engage you in the conversation (e.g., an opinion, thought)
            \begin{itemize}
                \item \textit{my favorite role would have to be quarterback . it is such an interesting role .}
                \item \textit{i love tom brady . i love tom brady .}
            \end{itemize}
            \item A score of 3 (interesting) means the response is very interesting or presents an interesting fact \begin{itemize}
                \item \textit{i agree . did you know that space jam is the highest grossing basketball movie of all time ?}
                \item \textit{a lot of players can make money by starring in movies . did you know space jam is the highest grossing movie of all time ? maybe one of the broke retired players can be in the sequel !}
            \end{itemize}            
        \end{itemize} \\ \hline

    \end{tabular}
    \caption{Annotation instructions (part 2 of 3)}
\end{table*}

\begin{table*}
    \begin{tabular}{|p{0.18\linewidth}|p{0.8\linewidth}|}
        \hline
        \multicolumn{2}{|c|}{\textbf{Annotation Instructions (ctd.)}} \\ \hline
  
        Uses Knowledge (0-1) &  Given the \textit{interesting fact} that the response is conditioned on, how well does the response use the fact?
        \begin{itemize}
            \item A score of 0 (no) means the response does not \textit{mention or refer to} the fact at all
            \item A score of 1 (yes) means the response uses the fact well     
        \end{itemize} \\ \hline

        Overall Quality (1-3) &   Given your answers above, what is your overall impression of this utterance?
        \begin{itemize}
            \item A score of 1 (very bad). A completely invalid response. It would be difficult to recover the conversation after this.
            \item A score of 2 (bad). Valid response, but otherwise poor in quality. 
            \item A score of 3 (neutral) means this response is neither good nor bad. This response has no negative qualities, but no positive ones either.
            \item A score of 4 (good) means this is a good response, but falls short of being perfect because of a key flaw.
            \item A score of 5 (very good) means this response is good and does not have any strong flaws.

        \end{itemize} \\ \hline

    \end{tabular}
    \caption{Annotation instructions (part 3 of 3)}
    
\end{table*}

\section{Metric Evaluation}

Table 3 in the main paper showed turn-level correlations for each specific quality. Due to space limitations, the table only included results for only the best correlated metrics. The full results are shown in Tables 9 - 21.

\section{Code and Data Release}

The code for the metrics can be found at \url{https://github.com/shikib/usr} and the human quality annotations can be found at \url{http://shikib.com/usr}. The human quality annotations will allow benchmarking of additional metrics. 

\begin{table*}
    \centering
    \renewcommand*{\arraystretch}{1.2}
    \begin{tabular}{|l|c|c|c|c|}
    \hline
        \textbf{Metric Name} & \multicolumn{2}{|c|}{\textbf{Turn-Level Correlation}} & \multicolumn{2}{|c|}{\textbf{System-Level Correlation}}  \\ \hline
         & Pearson & Spearman & Spearman & Pearson \\ \hline
         \multicolumn{5}{|c|}{Word-Overlap Metrics} \\ \hline
         F-1 & 0.1645 & 0.1690 & \textit{0.6000} & \textit{0.6120} \\
BLEU-1 & 0.2728 & 0.2876 & \textit{0.7000} & \textit{0.8334} \\
BLEU-2 & 0.2862 & 0.3012 & 0.9000 & \textit{0.8201} \\
BLEU-3 & 0.2569 & 0.3007 & 0.9000 & \textbf{0.9033} \\
BLEU-4 & 0.2160 & 0.2956 & 0.9000 & \textit{0.8740} \\
METEOR & 0.3365 & 0.3908 & 0.9000 & \textbf{0.9435} \\
ROUGE-L & 0.2745 & 0.2870 & 0.9000 & \textit{0.8143} \\ \hline
\multicolumn{5}{|c|}{Embedding-Based Metrics} \\ \hline
Greedy Matching & 0.1712 & 0.1943 & \textit{0.8000} & \textit{0.5610} \\
Embedding Average & 0.1803 & 0.2038 & \textit{0.7000} & \textbf{0.9166} \\
Vector Extrema & 0.2032 & 0.2091 & \textit{0.8000} & \textit{0.5838} \\
Skip-Thought & \textit{0.1040} & 0.1181 & \textit{0.5000} & \textit{0.5142} \\
BERTScore (base) & 0.3229 & 0.3540 & 0.9000 & \textbf{0.9100} \\
BERTScore (large) & 0.2982 & 0.3252 & 0.9000 & \textit{0.8536} \\ \hline
\multicolumn{5}{|c|}{Reference Free Metrics} \\ \hline
USR-MLM & 0.3086 & 0.3345 & 0.9000 & \textit{0.4732} \\
USR-DR (x = c) & 0.3245 & 0.4068 & \textit{0.7000} & \textbf{0.9182} \\
USR-DR (x = f) & 0.1419 & 0.3221 & 0.9000 & \textit{0.8519} \\
USR & \textbf{0.4192} & \textbf{0.4220} & \textbf{1.0000} & \textbf{0.9276} \\ \hline
    \end{tabular}
    \caption{Correlations of all the metrics with \textit{Overall Quality} ratings on Topical-Chat. All values with $p \geq 0.05$ are italicized.}
    
\end{table*}

\begin{table*}
    \centering
    \renewcommand*{\arraystretch}{1.2}
    \begin{tabular}{|l|c|c|c|c|}
    \hline
        \textbf{Metric Name} & \multicolumn{2}{|c|}{\textbf{Turn-Level Correlation}} & \multicolumn{2}{|c|}{\textbf{System-Level Correlation}}  \\ \hline
         & Pearson & Spearman & Spearman & Pearson \\ \hline
         \multicolumn{5}{|c|}{Word-Overlap Metrics} \\ \hline
F-1 & 0.1422 & \textit{0.1241} & 1.0000 & 0.9956 \\
BLEU-1 & \textit{0.0434} & \textit{0.0469} & \textit{0.6000} & \textit{0.2599} \\
BLEU-2 & \textit{0.1122} & \textit{0.0943} & \textit{0.4000} & \textit{0.6816} \\
BLEU-3 & \textit{0.1202} & \textit{0.0924} & \textit{0.4000} & \textit{0.6668} \\
BLEU-4 & 0.1353 & \textit{0.0899} & \textit{0.8000} & \textit{0.8413} \\
METEOR & 0.2527 & 0.2713 & \textit{0.8000} & \textit{0.9065} \\
ROUGE-L & \textit{0.0659} & \textit{0.0385} & \textit{0.0000} & \textit{0.1710} \\\hline 
 \multicolumn{5}{|c|}{Embedding-Based Metrics} \\ \hline
Greedy Matching & \textit{0.0916} & \textit{0.0625} & \textit{0.8000} & \textit{0.3808} \\
Embedding Average & \textit{0.1182} & 0.1428 & \textit{0.8000} & \textit{0.8628} \\
Vector Extrema & 0.1570 & 0.1410 & \textit{0.6000} & \textit{0.4349} \\
Skip-Thought & \textit{-0.0393} & \textit{-0.0452} & \textit{-0.2000} & \textit{0.2599} \\
BERTScore (base) & 0.1690 & 0.1526 & \textit{0.8000} & \textit{0.5173} \\
BERTScore (large) & 0.1518 & \textit{0.1211} & \textit{0.0000} & \textit{0.2410} \\\hline 
 \multicolumn{5}{|c|}{Reference Free Metrics} \\ \hline
USR-MLM & \textit{0.0795} & \textit{0.0788} & \textit{-0.4000} & \textit{-0.2842} \\
USR-DR (x = c) & \textbf{0.4814} & \textbf{0.6087} & \textbf{1.0000} & \textit{0.8202} \\
USR-DR (x = f) & \textit{-0.0495} & \textit{-0.0454} & \textit{-0.2108} & \textit{-0.0178} \\
USR & 0.4693 & 0.4115 & \textbf{1.0000} & \textit{0.8084} \\\hline
    \end{tabular}
    \caption{Correlations of all the metrics with \textit{Overall Quality} ratings on PersonaChat. All values with $p \geq 0.05$ are italicized.}
    
\end{table*}

\begin{table*}
    \centering
    \renewcommand*{\arraystretch}{1.2}
    \begin{tabular}{|l|c|c|c|c|}
    \hline
        \textbf{Metric Name} & \multicolumn{2}{|c|}{\textbf{Turn-Level Correlation}} & \multicolumn{2}{|c|}{\textbf{System-Level Correlation}}  \\ \hline
         & Pearson & Spearman & Spearman & Pearson \\ \hline
         \multicolumn{5}{|c|}{Word-Overlap Metrics} \\ \hline
F-1 & \textit{0.0425} & \textit{0.0620} & \textit{0.8000} & \textit{0.6481} \\
BLEU-1 & 0.1794 & 0.1522 & \textit{0.6000} & \textit{0.8360} \\
BLEU-2 & 0.2360 & 0.2081 & \textit{0.7000} & \textit{0.8262} \\
BLEU-3 & 0.2099 & 0.2137 & \textit{0.7000} & \textbf{0.9018} \\
BLEU-4 & 0.2010 & 0.2175 & \textit{0.7000} & \textit{0.8663} \\
METEOR & 0.2452 & 0.2246 & \textit{0.7000} & \textbf{0.9424} \\
ROUGE-L & 0.2069 & 0.1632 & \textit{0.7000} & \textit{0.8208} \\ \hline
\multicolumn{5}{|c|}{Embedding-Based Metrics} \\ \hline
Greedy Matching & \textit{0.0839} & \textit{0.0868} & \textit{0.6000} & \textit{0.5664} \\
Embedding Average & \textit{0.0509} & \textit{0.0961} & \textit{0.6000} & 0.9204 \\
Vector Extrema & 0.1561 & 0.1321 & \textit{0.6000} & \textit{0.6113} \\
Skip-Thought & \textit{0.0810} & \textit{0.0706} & \textit{0.2000} & \textit{0.4725} \\
BERTScore (base) & 0.2611 & 0.2502 & \textit{0.7000} & \textbf{0.9118} \\
BERTScore (large) & 0.2556 & 0.2263 & \textit{0.7000} & \textit{0.8577} \\ \hline
\multicolumn{5}{|c|}{Reference Free Metrics} \\ \hline
USR-MLM & \textbf{0.3264} & \textbf{0.3268} & \textit{0.7000} & \textit{0.4666} \\
USR-DR (x = c) & 0.1500 & 0.2213 & \textbf{0.9000} & \textbf{0.9337} \\
USR-DR (x = f) & \textit{0.0881} & 0.1967 & \textit{0.7000} & \textit{0.8420} \\
USR & 0.2932 & 0.3152 & \textbf{0.9000} & \textbf{0.9329} \\ \hline
    \end{tabular}
    \caption{Correlations of all the metrics with the \textit{Understandable} ratings on Topical-Chat. All values with $p \geq 0.05$ are italicized. The USR-MLM metric has poor system-level correlations, however the USR metric leverages predictions from the other sub-metrics to improve this.}
    
\end{table*}

\begin{table*}
    \centering
    \renewcommand*{\arraystretch}{1.2}
    \begin{tabular}{|l|c|c|c|c|}
    \hline
        \textbf{Metric Name} & \multicolumn{2}{|c|}{\textbf{Turn-Level Correlation}} & \multicolumn{2}{|c|}{\textbf{System-Level Correlation}}  \\ \hline
         & Pearson & Spearman & Spearman & Pearson \\ \hline
         \multicolumn{5}{|c|}{Word-Overlap Metrics} \\ \hline
F-1 & \textit{-0.0340} & \textit{-0.0550} & 1.0000 & 0.9956 \\
BLEU-1 & \textit{0.0123} & \textit{-0.0196} & \textit{0.6000} & \textit{0.2599} \\
BLEU-2 & \textit{0.0854} & \textit{0.0221} & \textit{0.4000} & \textit{0.6816} \\
BLEU-3 & \textit{0.0412} & \textit{0.0249} & \textit{0.4000} & \textit{0.6668} \\
BLEU-4 & \textit{0.0537} & \textit{0.0279} & \textit{0.8000} & \textit{0.8413} \\
METEOR & \textit{0.0820} & \textit{0.0431} & \textit{0.8000} & \textit{0.9065} \\
ROUGE-L & \textit{0.0346} & \textit{0.0076} & \textit{0.0000} & \textit{0.1710} \\\hline 
 \multicolumn{5}{|c|}{Embedding-Based Metrics} \\ \hline
Greedy Matching & \textit{0.0594} & \textit{0.0710} & \textit{0.8000} & \textit{0.3808} \\
Embedding Average & \textit{0.0573} & \textit{0.0835} & \textit{0.8000} & \textit{0.8628} \\
Vector Extrema & \textit{0.1097} & \textit{0.1113} & \textit{0.6000} & \textit{0.4349} \\
Skip-Thought & \textit{-0.0338} & \textit{-0.0297} & \textit{-0.2000} & \textit{0.2599} \\
BERTScore (base) & \textit{0.0676} & \textit{0.0685} & \textit{0.8000} & \textit{0.5173} \\
BERTScore (large) & \textit{0.0380} & \textit{0.0086} & \textit{0.0000} & \textit{0.2410} \\\hline 
 \multicolumn{5}{|c|}{Reference Free Metrics} \\ \hline
USR-MLM & \textbf{0.1313} & \textit{0.1186} & \textit{-0.4000} & \textit{-0.2842} \\
USR-DR (x = c) & \textit{0.0728} & \textbf{0.1446} & \textbf{1.0000} & \textit{0.8202} \\
USR-DR (x = f) & \textit{-0.0390} & \textit{-0.0433} & \textit{-0.2108} & \textit{-0.0178} \\
USR & \textit{0.0997} & 0.1337 & \textbf{1.0000} & \textit{0.8084} \\\hline
    \end{tabular}
    \caption{Correlations of all the metrics with \textit{Understandable} ratings on PersonaChat. All values with $p \geq 0.05$ are italicized.}
    
\end{table*}

\begin{table*}
    \centering
    \renewcommand*{\arraystretch}{1.2}
    \begin{tabular}{|l|c|c|c|c|}
    \hline
        \textbf{Metric Name} & \multicolumn{2}{|c|}{\textbf{Turn-Level Correlation}} & \multicolumn{2}{|c|}{\textbf{System-Level Correlation}}  \\ \hline
         & Pearson & Spearman & Spearman & Pearson \\ \hline
         \multicolumn{5}{|c|}{Word-Overlap Metrics} \\ \hline
F-1 & \textit{0.0301} & \textit{0.0398} & \textit{0.6000} & \textit{0.5605} \\
BLEU-1 & 0.1606 & 0.1334 & \textit{0.7000} & \textit{0.7976} \\
BLEU-2 & 0.1959 & 0.1648 & 0.9000 & \textit{0.7888} \\
BLEU-3 & 0.1896 & 0.1745 & 0.9000 & 0.8979 \\
BLEU-4 & 0.1799 & 0.1748 & 0.9000 & 0.8973 \\
METEOR & 0.2121 & 0.1906 & 0.9000 & \textbf{0.9297} \\
ROUGE-L & 0.1760 & 0.1457 & 0.9000 & \textit{0.7902} \\  \hline
\multicolumn{5}{|c|}{Embedding-Based Metrics} \\ \hline
Greedy Matching & \textit{0.0534} & \textit{0.0483} & \textit{0.8000} & \textit{0.5271} \\
Embedding Average & \textit{0.0477} & \textit{0.0970} & \textit{0.7000} & 0.8875 \\
Vector Extrema & \textit{0.1009} & \textit{0.0761} & \textit{0.8000} & \textit{0.5363} \\
Skip-Thought & \textit{0.0959} & \textit{0.0858} & \textit{0.5000} & \textit{0.5313} \\
BERTScore (base) & 0.2164 & 0.2088 & 0.9000 & \textbf{0.9024} \\
BERTScore (large) & 0.2260 & 0.2094 & 0.9000 & \textit{0.8319} \\ \hline
\multicolumn{5}{|c|}{Reference Free Metrics} \\ \hline
USR-MLM & \textbf{0.3370} & \textbf{0.3254} & 0.9000 & \textit{0.4485} \\
USR-DR (x = c) & 0.1325 & 0.2148 & \textit{0.7000} & \textbf{0.9222} \\
USR-DR (x = f) & \textit{0.0313} & 0.1611 & 0.9000 & 0.8808 \\
USR & 0.2763 & 0.3037 & \textbf{1.0000} & \textbf{0.9220} \\ \hline
    \end{tabular}
    \caption{Correlations of all the metrics with the \textit{Natural} ratings on Topical-Chat. All values with $p \geq 0.05$ are italicized. The USR-MLM metric has poor system-level correlations, however the USR metric leverages predictions from the other sub-metrics to improve this. }
    
\end{table*}

\begin{table*}
    \centering
    \renewcommand*{\arraystretch}{1.2}
    \begin{tabular}{|l|c|c|c|c|}
    \hline
        \textbf{Metric Name} & \multicolumn{2}{|c|}{\textbf{Turn-Level Correlation}} & \multicolumn{2}{|c|}{\textbf{System-Level Correlation}}  \\ \hline
         & Pearson & Spearman & Spearman & Pearson \\ \hline
         \multicolumn{5}{|c|}{Word-Overlap Metrics} \\ \hline
F-1 & \textit{0.0815} & \textit{0.0717} & 1.0000 & 0.9956 \\
BLEU-1 & \textit{-0.0072} & \textit{-0.0216} & \textit{0.6000} & \textit{0.2599} \\
BLEU-2 & \textit{0.0838} & \textit{0.0344} & \textit{0.4000} & \textit{0.6816} \\
BLEU-3 & \textit{0.0823} & \textit{0.0457} & \textit{0.4000} & \textit{0.6668} \\
BLEU-4 & \textit{0.1081} & \textit{0.0499} & \textit{0.8000} & \textit{0.8413} \\
METEOR & \textit{0.0989} & \textit{0.0950} & \textit{0.8000} & \textit{0.9065} \\
ROUGE-L & \textit{0.0096} & \textit{-0.0087} & \textit{0.0000} & \textit{0.1710} \\\hline 
 \multicolumn{5}{|c|}{Embedding-Based Metrics} \\ \hline
Greedy Matching & \textit{0.1029} & \textit{0.0665} & \textit{0.8000} & \textit{0.3808} \\
Embedding Average & 0.1413 & \textit{0.1152} & \textit{0.8000} & \textit{0.8628} \\
Vector Extrema & 0.1458 & 0.1375 & \textit{0.6000} & \textit{0.4349} \\
Skip-Thought & \textit{-0.0355} & \textit{-0.0365} & \textit{-0.2000} & \textit{0.2599} \\
BERTScore (base) & \textit{0.0606} & \textit{0.0585} & \textit{0.8000} & \textit{0.5173} \\
BERTScore (large) & \textit{0.0494} & \textit{0.0477} & \textit{0.0000} & \textit{0.2410} \\\hline 
 \multicolumn{5}{|c|}{Reference Free Metrics} \\ \hline
USR-MLM & \textit{0.0999} & \textit{0.1119} & \textit{-0.4000} & \textit{-0.2842} \\
USR-DR (x = c) & 0.1733 & 0.2291 & \textbf{1.0000} & \textit{0.8202} \\
USR-DR (x = f) & \textit{-0.0033} & \textit{0.0642} & \textit{-0.2108} & \textit{-0.0178} \\
USR & \textbf{0.1862} & \textbf{0.2430} & \textbf{1.0000} & \textit{0.8084} \\ \hline
    \end{tabular}
    \caption{Correlations of all the metrics with the \textit{Natural} ratings on PersonaChat. All values with $p \geq 0.05$ are italicized. }
    
\end{table*}

\begin{table*}
    \centering
    \renewcommand*{\arraystretch}{1.2}
    \begin{tabular}{|l|c|c|c|c|}
    \hline
        \textbf{Metric Name} & \multicolumn{2}{|c|}{\textbf{Turn-Level Correlation}} & \multicolumn{2}{|c|}{\textbf{System-Level Correlation}}  \\ \hline
         & Pearson & Spearman & Spearman & Pearson \\ \hline
         \multicolumn{5}{|c|}{Word-Overlap Metrics} \\ \hline
F-1 & 0.1290 & 0.1199 & \textit{0.6000} & \textit{0.6483} \\
BLEU-1 & 0.2097 & 0.2228 & \textbf{1.0000} & \textit{0.8754} \\
BLEU-2 & 0.2087 & 0.2353 & 0.9000 & \textit{0.8555} \\
BLEU-3 & 0.1736 & 0.2377 & 0.9000 & \textbf{0.9090} \\
BLEU-4 & 0.1307 & 0.2345 & \textit{0.5000} & \textit{0.8464} \\
METEOR & 0.2495 & 0.3018 & 0.9000 & \textbf{0.9573} \\
ROUGE-L & 0.1928 & 0.2031 & 0.9000 & \textit{0.8410} \\   \hline
\multicolumn{5}{|c|}{Embedding-Based Metrics} \\ \hline
Greedy Matching & \textit{0.1036} & 0.1249 & \textit{0.8000} & \textit{0.6078} \\
Embedding Average & 0.1197 & 0.1511 & \textbf{1.0000} & \textbf{0.9460} \\
Vector Extrema & 0.1839 & 0.1840 & \textit{0.8000} & \textit{0.6275} \\
Skip-Thought & \textit{0.0326} & \textit{0.0568} & \textit{0.6000} & \textit{0.5237} \\
BERTScore (base) & 0.2432 & 0.2642 & 0.9000 & \textbf{0.9160} \\
BERTScore (large) & 0.2140 & 0.2328 & 0.9000 & \textit{0.8779} \\ \hline
\multicolumn{5}{|c|}{Reference Free Metrics} \\ \hline
USR-MLM & 0.3099 & 0.3243 & 0.9000 & \textit{0.5190} \\
USR-DR (x = c) & 0.3391 & \textbf{0.3650} & \textit{0.3000} & 0.8899 \\
USR-DR (x = f) & \textit{0.0594} & 0.1836 & \textit{0.5000} & \textit{0.8188} \\
USR & \textbf{0.4160} & \textbf{0.3769} & \textit{0.7000} & \textbf{0.9270} \\ \hline
    \end{tabular}
    \caption{Correlations of all the metrics with the \textit{Maintains Context} ratings on Topical-Chat. All values with $p \geq 0.05$ are italicized. Several referenced metrics perform strongly on the system-level correlations, however USR strongly outperforms all other metrics on the turn-level correlations.}
    
\end{table*}

\begin{table*}
    \centering
    \renewcommand*{\arraystretch}{1.2}
    \begin{tabular}{|l|c|c|c|c|}
    \hline
        \textbf{Metric Name} & \multicolumn{2}{|c|}{\textbf{Turn-Level Correlation}} & \multicolumn{2}{|c|}{\textbf{System-Level Correlation}}  \\ \hline
         & Pearson & Spearman & Spearman & Pearson \\ \hline
         \multicolumn{5}{|c|}{Word-Overlap Metrics} \\ \hline
F-1 & \textit{0.1073} & \textit{0.0747} & 1.0000 & 0.9956 \\
BLEU-1 & \textit{0.0713} & \textit{0.0799} & \textit{0.6000} & \textit{0.2599} \\
BLEU-2 & \textit{0.0949} & 0.1372 & \textit{0.4000} & \textit{0.6816} \\
BLEU-3 & 0.1270 & 0.1461 & \textit{0.4000} & \textit{0.6668} \\
BLEU-4 & 0.1467 & 0.1508 & \textit{0.8000} & \textit{0.8413} \\
METEOR & 0.2500 & 0.2564 & \textit{0.8000} & \textit{0.9065} \\
ROUGE-L & \textit{0.1135} & \textit{0.0910} & \textit{0.0000} & \textit{0.1710} \\\hline 
 \multicolumn{5}{|c|}{Embedding-Based Metrics} \\ \hline
Greedy Matching & 0.1503 & 0.1631 & \textit{0.8000} & \textit{0.3808} \\
Embedding Average & \textit{0.1010} & 0.1660 & \textit{0.8000} & \textit{0.8628} \\
Vector Extrema & 0.2288 & 0.2631 & \textit{0.6000} & \textit{0.4349} \\
Skip-Thought & \textit{0.0243} & \textit{0.0139} & \textit{-0.2000} & \textit{0.2599} \\
BERTScore (base) & 0.1770 & 0.1686 & \textit{0.8000} & \textit{0.5173} \\
BERTScore (large) & 0.1877 & 0.1569 & \textit{0.0000} & \textit{0.2410} \\\hline 
 \multicolumn{5}{|c|}{Reference Free Metrics} \\ \hline
USR-MLM & 0.1805 & 0.2067 & \textit{-0.4000} & \textit{-0.2842} \\
USR-DR (x = c) & 0.6021 & \textbf{0.5625} & \textbf{1.0000} & \textit{0.8202} \\
USR-DR (x = f) & \textit{-0.0198} & \textit{-0.0164} & \textit{-0.2108} & \textit{-0.0178} \\
USR & \textbf{0.6065} & 0.5280 & \textbf{1.0000} & \textit{0.8084} \\ \hline
    \end{tabular}
    \caption{Correlations of all the metrics with the \textit{Maintains Context} ratings on PersonaChat. All values with $p \geq 0.05$ are italicized. Several referenced metrics perform strongly on the system-level correlations, however USR strongly outperforms all other metrics on the turn-level correlations.}
    
\end{table*}

\begin{table*}
    \centering
    \renewcommand*{\arraystretch}{1.2}
    \begin{tabular}{|l|c|c|c|c|}
    \hline
        \textbf{Metric Name} & \multicolumn{2}{|c|}{\textbf{Turn-Level Correlation}} & \multicolumn{2}{|c|}{\textbf{System-Level Correlation}}  \\ \hline
         & Pearson & Spearman & Spearman & Pearson \\ \hline
         \multicolumn{5}{|c|}{Word-Overlap Metrics} \\ \hline
F-1 & 0.2523 & 0.2565 & \textit{0.6000} & \textit{0.5944} \\
BLEU-1 & 0.3144 & 0.3343 & \textit{0.7000} & \textit{0.8197} \\
BLEU-2 & 0.3184 & 0.3323 & 0.9000 & \textit{0.8099} \\
BLEU-3 & 0.2782 & 0.3247 & 0.9000 & \textbf{0.9047} \\
BLEU-4 & 0.2322 & 0.3156 & 0.9000 & 0.8883 \\
METEOR & 0.3668 & 0.4391 & 0.9000 & \textbf{0.9398} \\
ROUGE-L & 0.2946 & 0.2995 & 0.9000 & \textit{0.8084} \\ \hline
\multicolumn{5}{|c|}{Embedding-Based Metrics} \\ \hline
Greedy Matching & 0.1989 & 0.2111 & \textit{0.8000} & \textit{0.5512} \\
Embedding Average & 0.1940 & 0.2161 & \textit{0.7000} & \textbf{0.9056} \\
Vector Extrema & 0.2101 & 0.2050 & \textit{0.8000} & \textit{0.5694} \\
Skip-Thought & 0.1139 & 0.1356 & \textit{0.5000} & \textit{0.5187} \\
BERTScore (base) & 0.3512 & 0.3725 & 0.9000 & \textbf{0.9108} \\
BERTScore (large) & 0.3167 & 0.3349 & 0.9000 & \textit{0.8480} \\ \hline
\multicolumn{5}{|c|}{Reference Free Metrics} \\ \hline
USR-MLM & 0.3189 & 0.3337 & 0.9000 & \textit{0.4663} \\
USR-DR (x = c) & 0.3533 & \textbf{0.4877} & \textit{0.7000} & \textbf{0.9233} \\
USR-DR (x = f) & 0.2006 & 0.4110 & 0.9000 & \textit{0.8685} \\
USR & \textbf{0.4555} & 0.4645 & \textbf{1.0000} & \textbf{0.9297} \\  \hline
    \end{tabular}
    \caption{Correlations of all the metrics with the \textit{Interesting} ratings on Topical-Chat. All values with $p \geq 0.05$ are italicized.}
    
\end{table*}

\begin{table*}
    \centering
    \renewcommand*{\arraystretch}{1.2}
    \begin{tabular}{|l|c|c|c|c|}
    \hline
        \textbf{Metric Name} & \multicolumn{2}{|c|}{\textbf{Turn-Level Correlation}} & \multicolumn{2}{|c|}{\textbf{System-Level Correlation}}  \\ \hline
         & Pearson & Spearman & Spearman & Pearson \\ \hline
         \multicolumn{5}{|c|}{Word-Overlap Metrics} \\ \hline
F-1 & \textit{0.0473} & \textit{0.0132} & 1.0000 & 0.9956 \\
BLEU-1 & \textit{-0.1081} & \textit{-0.0922} & \textit{0.6000} & \textit{0.2599} \\
BLEU-2 & \textit{-0.1048} & \textit{-0.1010} & \textit{0.4000} & \textit{0.6816} \\
BLEU-3 & \textit{-0.1247} & \textit{-0.1151} & \textit{0.4000} & \textit{0.6668} \\
BLEU-4 & -0.1359 & \textit{-0.1242} & \textit{0.8000} & \textit{0.8413} \\
METEOR & \textit{-0.0458} & \textit{0.0116} & \textit{0.8000} & \textit{0.9065} \\
ROUGE-L & -0.1456 & -0.1354 & \textit{0.0000} & \textit{0.1710} \\\hline 
 \multicolumn{5}{|c|}{Embedding-Based Metrics} \\ \hline
Greedy Matching & -0.1778 & -0.2080 & \textit{0.8000} & \textit{0.3808} \\
Embedding Average & \textit{-0.0141} & \textit{-0.0177} & \textit{0.8000} & \textit{0.8628} \\
Vector Extrema & -0.1883 & -0.1746 & \textit{0.6000} & \textit{0.4349} \\
Skip-Thought & \textit{-0.0882} & \textit{-0.0916} & \textit{-0.2000} & \textit{0.2599} \\
BERTScore (base) & \textit{0.0325} & \textit{0.0491} & \textit{0.8000} & \textit{0.5173} \\
BERTScore (large) & \textit{-0.0418} & \textit{-0.0245} & \textit{0.0000} & \textit{0.2410} \\\hline 
 \multicolumn{5}{|c|}{Reference Free Metrics} \\ \hline
USR-MLM & \textit{-0.1045} & \textit{-0.1007} & \textit{-0.4000} & \textit{-0.2842} \\
USR-DR (x = c) & \textit{0.0606} & \textbf{0.2634} & \textbf{1.0000} & \textit{0.8202} \\
USR-DR (x = f) & \textit{-0.0022} & \textit{-0.0039} & \textit{-0.2108} & \textit{-0.0178} \\
USR & \textit{0.0315} & \textit{0.0171} & \textbf{1.0000} & \textit{0.8084} \\  \hline
    \end{tabular}
    \caption{Correlations of all the metrics with the \textit{Interesting} ratings on PersonaChat. All values with $p \geq 0.05$ are italicized.}
    
\end{table*}

\begin{table*}
    \centering
    \renewcommand*{\arraystretch}{1.2}
    \begin{tabular}{|l|c|c|c|c|}
    \hline
        \textbf{Metric Name} & \multicolumn{2}{|c|}{\textbf{Turn-Level Correlation}} & \multicolumn{2}{|c|}{\textbf{System-Level Correlation}}  \\ \hline
         & Pearson & Spearman & Spearman & Pearson \\ \hline
         \multicolumn{5}{|c|}{Word-Overlap Metrics} \\ \hline
F-1 & 0.1495 & 0.1485 & \textit{0.6000} & \textit{0.5970} \\
BLEU-1 & 0.2888 & 0.3033 & \textit{0.7000} & \textit{0.8357} \\
BLEU-2 & 0.2819 & 0.3066 & 0.9000 & \textit{0.8309} \\
BLEU-3 & 0.2442 & 0.3106 & 0.9000 & \textbf{0.9259} \\
BLEU-4 & 0.2126 & 0.3096 & 0.9000 & \textbf{0.9084} \\
METEOR & 0.3328 & 0.3909 & 0.9000 & \textbf{0.9534} \\
ROUGE-L & 0.3099 & 0.3273 & 0.9000 & \textit{0.8333} \\  \hline
\multicolumn{5}{|c|}{Embedding-Based Metrics} \\ \hline
Greedy Matching & 0.2327 & 0.2306 & \textit{0.8000} & \textit{0.5874} \\
Embedding Average & 0.1812 & 0.1827 & \textit{0.7000} & \textbf{0.9151} \\
Vector Extrema & 0.2294 & 0.2111 & \textit{0.8000} & \textit{0.5917} \\
Skip-Thought & \textit{0.0986} & 0.1145 & \textit{0.5000} & \textit{0.5397} \\
BERTScore (base) & 0.2847 & 0.2947 & 0.9000 & \textbf{0.9308} \\
BERTScore (large) & 0.2909 & 0.3167 & 0.9000 & \textit{0.8703} \\ \hline
\multicolumn{5}{|c|}{Reference Free Metrics} \\ \hline
USR-MLM & 0.2195 & 0.2261 & 0.9000 & \textit{0.5070} \\
USR-DR (x = c) & 0.2285 & 0.4179 & \textit{0.7000} & \textbf{0.9155} \\
USR-DR (x = f) & 0.2220 & \textbf{0.4468} & 0.9000 & 0.8884 \\
USR & \textbf{0.3175} & 0.3353 & \textbf{1.0000} & \textbf{0.9469} \\ \hline
    \end{tabular}
    \caption{Correlations of all the metrics with the \textit{Uses Knowledge} ratings on Topical-Chat. All values with $p \geq 0.05$ are italicized.}
    
\end{table*}

\begin{table*}
    \centering
    \renewcommand*{\arraystretch}{1.2}
    \begin{tabular}{|l|c|c|c|c|}
    \hline
        \textbf{Metric Name} & \multicolumn{2}{|c|}{\textbf{Turn-Level Correlation}} & \multicolumn{2}{|c|}{\textbf{System-Level Correlation}}  \\ \hline
         & Pearson & Spearman & Spearman & Pearson \\ \hline
         \multicolumn{5}{|c|}{Word-Overlap Metrics} \\ \hline
F-1 & \textit{0.0869} & \textit{0.1056} & 1.0000 & 0.9956 \\
BLEU-1 & \textit{0.0737} & \textit{0.0729} & \textit{0.6000} & \textit{0.2599} \\
BLEU-2 & \textit{0.1083} & \textit{0.0722} & \textit{0.4000} & \textit{0.6816} \\
BLEU-3 & \textit{0.0999} & \textit{0.0594} & \textit{0.4000} & \textit{0.6668} \\
BLEU-4 & \textit{0.0698} & \textit{0.0528} & \textit{0.8000} & \textit{0.8413} \\
METEOR & 0.1678 & 0.1719 & \textit{0.8000} & \textit{0.9065} \\
ROUGE-L & \textit{0.0710} & \textit{0.0632} & \textit{0.0000} & \textit{0.1710} \\\hline 
 \multicolumn{5}{|c|}{Embedding-Based Metrics} \\ \hline
Greedy Matching & \textit{0.0382} & \textit{0.0057} & \textit{0.8000} & \textit{0.3808} \\
Embedding Average & \textit{0.0402} & \textit{0.0618} & \textit{0.8000} & \textit{0.8628} \\
Vector Extrema & \textit{0.0564} & \textit{-0.0008} & \textit{0.6000} & \textit{0.4349} \\
Skip-Thought & \textit{-0.0686} & \textit{-0.0609} & \textit{-0.2000} & \textit{0.2599} \\
BERTScore (base) & \textit{0.0719} & \textit{0.0465} & \textit{0.8000} & \textit{0.5173} \\
BERTScore (large) & \textit{0.0271} & \textit{0.0094} & \textit{0.0000} & \textit{0.2410} \\\hline 
 \multicolumn{5}{|c|}{Reference Free Metrics} \\ \hline
USR-MLM & \textit{-0.0782} & \textit{-0.0756} & \textit{-0.4000} & \textit{-0.2842} \\
USR-DR (x = c) & \textbf{0.4508} & \textbf{0.6309} & \textbf{1.0000} & \textit{0.8202} \\
USR-DR (x = f) & \textit{-0.0927} & \textit{-0.0903} & \textit{-0.2108} & \textit{-0.0178} \\
USR & 0.4027 & 0.3177 & \textbf{1.0000} & \textit{0.8084} \\ \hline
    \end{tabular}
    \caption{Correlations of all the metrics with the \textit{Uses Knowledge} ratings on PersonaChat. All values with $p \geq 0.05$ are italicized.}
    
\end{table*}

\end{document}